\documentclass{article}

\PassOptionsToPackage{numbers,sort&compress}{natbib}
\usepackage[dblblindworkshop, final]{neurips_2026}  

\workshoptitle{TAE (Trust-AI-Eval): Can We Trust AI Evaluation?}

\usepackage[utf8]{inputenc} 
\usepackage[T1]{fontenc}    
\usepackage{hyperref}       
\usepackage{url}            
\usepackage{booktabs}       
\usepackage{amsfonts}       
\usepackage{nicefrac}       
\usepackage{microtype}      
\usepackage{xcolor}         

\usepackage{amsmath}
\usepackage{amssymb}
\usepackage{graphicx}
\usepackage{tabularx}
\usepackage{placeins}
\usepackage{multirow}
\usepackage{makecell}
\usepackage{subcaption}
\usepackage{enumitem}
\usepackage{tcolorbox}
\tcbuselibrary{breakable, skins}
\usepackage{tikz}
\usetikzlibrary{arrows.meta, positioning}

\setcitestyle{numbers,square,comma,sort&compress}

\definecolor{darkblue}{rgb}{0, 0, 0.5}
\hypersetup{colorlinks=true, citecolor=darkblue, linkcolor=darkblue, urlcolor=darkblue}

\title{Evolving LLM-Generated Features for Interpretable Classification}

\author{%
  Jack Butler \\
  Amazon Web Services \\
  \texttt{jackbtlr@amazon.com} \\
  \And
  Zainab Afolabi \\
  Amazon Web Services \\
  \texttt{zafolabi@amazon.com} \\
  \And
  Nikita Kozodoi \\
  Amazon Web Services \\
  \texttt{kozodoi@amazon.com} \\
}

\begin{document}

\maketitle

\begin{abstract}
Large language models (LLMs) are increasingly used as classifiers, yet they operate as opaque systems whose decisions are difficult to interpret, which complicates their use in regulated domains such as credit scoring or medical diagnosis. We propose an evolutionary framework that iteratively discovers natural language feature definitions (rubrics) for interpretable classification. An LLM generates candidate binary features, evaluates each sample against them, and the resulting vectors can be used to train a transparent classifier such as logistic regression. The feature set evolves over multiple iterations guided by classification errors, per-class activation rates, and feature ablation scores. We evaluate across three benchmarks, including a credit risk dataset representative of regulated domains, comparing single-shot LLM rubrics, evolved rubrics, and direct zero-shot LLM classification. Evolved features improve over single-shot rubrics by +2.9 pp on average and outperform zero-shot LLM classification on two of three tasks, while providing fully auditable decision logic. On the credit risk task, the zero-shot LLM performs at chance (50.7\%) with a strong bias toward a single class, whereas evolved features achieve balanced, interpretable predictions. Crucially, this failure is invisible in aggregate accuracy and surfaces only under per-class auditing. Analysis reveals that evolution is most effective when label boundaries cannot be inferred from category names alone or when the LLM lacks reliable domain-specific reasoning.
\end{abstract}

%
%

\section{Introduction}
\label{sec_introduction}

Large Language Models (LLMs) demonstrate strong performance across diverse classification tasks, from sentiment analysis to content moderation~\citep{brown2020language, wei2022chain, zheng2023judging}. However, these models operate as opaque systems whose internal decision processes are difficult to interpret or audit. In high-stakes settings such as credit scoring, medical diagnosis, and content moderation, practitioners require not only accurate predictions but also transparent explanations of the factors driving them~\citep{rudin2019stop}. This is acute in consumer lending, where decisions must carry adverse-action explanations under the Equal Credit Opportunity Act, and a classifier can fail silently by defaulting to one class while aggregate accuracy still looks plausible.

Traditional interpretable models (linear classifiers, decision trees, and rule-based systems) provide transparent decision processes but rely on manually engineered features that may fail to capture complex semantic patterns in natural language. Recent work has explored using LLMs to generate interpretable features~\citep{singh2023augmenting, balek2024llms}. However, these approaches typically produce static feature sets that do not adapt to dataset-specific patterns or classification errors observed during training.

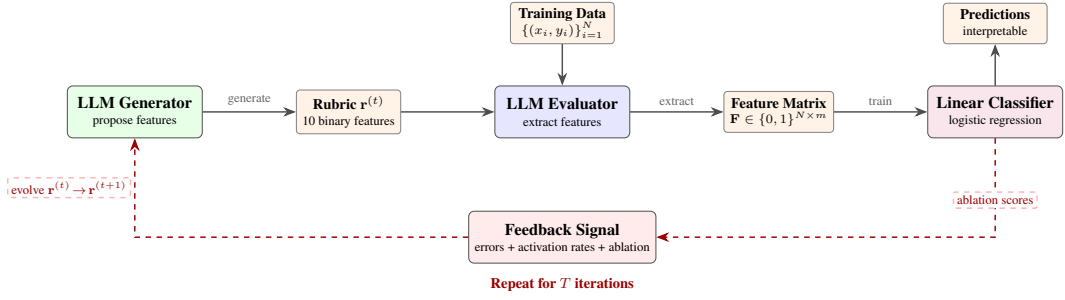
\begin{figure*}[t]
\centering
\resizebox{\textwidth}{!}{%
\begin{tikzpicture}[
    node distance=1.4cm and 2.0cm,
    box/.style={rectangle, draw=black!70, fill=#1, rounded corners=3pt,
                minimum height=1.0cm, minimum width=2.6cm, align=center,
                font=\small},
    box/.default={blue!8},
    data/.style={rectangle, draw=black!50, fill=orange!10, rounded corners=2pt,
                 minimum height=0.7cm, minimum width=2.0cm, align=center,
                 font=\footnotesize},
    arr/.style={-{Stealth[length=2.5mm]}, thick, black!70},
    feedback/.style={-{Stealth[length=2.5mm]}, thick, red!60!black, dashed},
    lbl/.style={font=\scriptsize, text=black!60},
    lblbox/.style={font=\scriptsize, text=red!60!black, fill=white, draw=red!30, rounded corners=1pt, inner sep=2pt},
]

\node[box=green!10] (gen) {\textbf{LLM Generator}\\[-1pt]{\scriptsize propose features}};
\node[data, right=1.8cm of gen] (rubric) {\textbf{Rubric} $\mathbf{r}^{(t)}$\\[-1pt]{\scriptsize 10 binary features}};
\node[box=blue!10, right=1.8cm of rubric] (eval) {\textbf{LLM Evaluator}\\[-1pt]{\scriptsize extract features}};
\node[data, right=1.8cm of eval] (matrix) {\textbf{Feature Matrix}\\[-1pt]{\scriptsize $\mathbf{F} \in \{0,1\}^{N \times m}$}};
\node[box=purple!10, right=1.8cm of matrix] (clf) {\textbf{Linear Classifier}\\[-1pt]{\scriptsize logistic regression}};

\node[data, above=0.8cm of clf] (pred) {\textbf{Predictions}\\[-1pt]{\scriptsize interpretable}};

\node[box=red!8, below=1.4cm of eval] (feedback) {\textbf{Feedback Signal}\\[-1pt]{\scriptsize errors + activation rates + ablation}};

\node[data, above=0.8cm of eval] (traindata) {\textbf{Training Data}\\[-1pt]{\scriptsize $\{(x_i, y_i)\}_{i=1}^{N}$}};

\draw[arr] (gen) -- (rubric) node[lbl, midway, above] {generate};
\draw[arr] (rubric) -- (eval) node[lbl, midway, above] {};
\draw[arr] (eval) -- (matrix) node[lbl, midway, above] {extract};
\draw[arr] (matrix) -- (clf) node[lbl, midway, above] {train};
\draw[arr] (clf) -- (pred);
\draw[arr] (traindata) -- (eval);

\draw[feedback] (clf.south) |- (feedback.east)
    node[lblbox, pos=0.25, below=2pt] {ablation scores};
\draw[feedback] (feedback.west) -| (gen.south)
    node[lblbox, pos=0.75, left=2pt] {evolve $\mathbf{r}^{(t)} \!\to\! \mathbf{r}^{(t+1)}$};

\node[font=\footnotesize\bfseries, text=red!60!black, below=0.15cm of feedback]
    {Repeat for $T$ iterations};

\end{tikzpicture}%
}
\caption{Overview of the evolutionary feature discovery framework. An LLM generates binary feature definitions (the rubric), evaluates each training sample against them, and a logistic regression classifier is trained on the resulting feature matrix. Classification errors, per-class activation rates, and feature ablation scores are fed back to guide the next rubric proposal (dashed arrows).}
\label{fig:framework}
\end{figure*}

We propose an evolutionary framework that bridges this gap by combining the semantic understanding of LLMs with the transparency of linear classifiers. Our approach uses an LLM to iteratively generate and refine binary features (rubrics), i.e., short, human-readable descriptions of binary attributes. An LLM evaluates these features on each sample, producing feature vectors that train a logistic regression classifier. The feature set evolves over multiple generations, with performance and feature importance guiding which features are retained, modified, or created (Figure~\ref{fig:framework}).

On a news task, the framework discovers features such as ``Contains sports-specific terminology (athlete names, team abbreviations, league names)'' ($|\bar{w}|=0.85$) that a practitioner can validate.

We make two contributions. First, we introduce an evolutionary framework that iteratively discovers natural language feature definitions from LLM-generated hypotheses and downstream performance feedback, and we compare it against single-shot LLM rubrics and zero-shot LLM classification across three benchmarks spanning text classification and credit risk assessment. Second, we show that aggregate accuracy alone can hide a systematic single-class bias that drives a classifier to chance-level performance, which an interpretable feature layer exposes and which we characterise in terms of label taxonomy and LLM domain competence.

%
%

\section{Related work}
\label{sec_related_work}

LLMs can perform classification through prompting, often achieving strong few-shot performance~\citep{brown2020language, wei2022chain, zheng2023judging}. Key advantages include flexibility when class definitions change (no retraining required) and the ability to classify without labelled training data. However, these approaches rely on implicit reasoning within the model, making predictions difficult to inspect or audit. Even chain-of-thought prompting~\citep{wei2022chain}, while producing intermediate reasoning steps, does not guarantee that the stated rationale reflects the model's actual decision process.

Traditional interpretable models (linear classifiers, decision trees, and rule-based systems) provide transparent decision processes through explicitly defined features~\citep{rudin2019stop}. Concept Bottleneck Models (CBMs) formalize this by routing predictions through human-readable concept layers~\citep{yang2023language, sun2024concept}, and textual CBMs build such concept bases for text classification without predefined human-labelled concepts~\citep{bhan2025completeness}. Recent work on explainability for LLMs (e.g., attention visualisation, mechanistic interpretability) probes model internals, but these post-hoc methods remain far from the transparency of coefficient inspection in linear models, which regulators in finance and healthcare already accept as sufficient explanation~\citep{rudin2019stop}. We extend the concept bottleneck paradigm to text classification with automatically discovered concepts.

\citet{singh2023augmenting} demonstrate that LLMs can augment interpretable models by generating natural language explanations that serve as features. \citet{balek2024llms} propose using LLMs as interpretable feature generators for classification, showing that LLM-generated features can match or exceed hand-crafted ones. However, both produce static feature sets generated once at the start, which never adapt to error patterns observed during training.

OPRO~\citep{yang2024opro} demonstrates that LLMs can serve as optimisers when provided with solution-score pairs, iteratively improving prompts through natural language feedback. \citet{videau2024evolutionary} apply evolutionary algorithms to optimise few-shot prompt selection. These methods optimise the prompt given to an LLM classifier, whereas we optimise the features used by a separate transparent classifier, a different decomposition that preserves interpretability.

Genetic programming has long been used to discover features or model structures through iterative modification based on performance feedback~\citep{koza1992genetic}. More recently, \citet{chen2023evoprompting} combine LLMs with evolutionary search to discover neural architectures, and AlphaEvolve~\citep{novikov2025alphaevolve} pairs an LLM mutation operator with automated evaluators to discover novel algorithms, underscoring the power of LLM-driven evolutionary search. Closest to our setting, LLM-FE~\citep{abhyankar2025llmfe} and CAAFE~\citep{hollmann2023caafe} have an LLM iteratively generate tabular features as executable code guided by validation feedback. The distinction is what the LLM writes: these methods emit transformations over existing columns, whereas our rubrics are natural language predicates an LLM evaluates against raw text, which applies where there are no columns to transform and keeps each feature legible to a non-programmer. Concurrent to our work, AutoQual~\citep{lan2025autoqual} also discovers interpretable natural-language features for a linear predictor, but selects them by mutual-information beam search for review-quality regression; we instead drive discovery with classification errors, per-class activation rates, and ablation scores, and use the resulting transparency to expose a per-class failure mode. Our framework is thus evolutionary search where the LLM is both mutation operator and fitness evaluator over human-readable descriptions rather than code.

The LLM-as-judge paradigm~\citep{zheng2023judging} has shown that LLMs can reliably evaluate complex outputs, achieving agreement rates comparable to human annotators. Our framework uses this capability: the evaluator LLM acts as a structured judge, assessing whether each binary feature is present in a sample and returning parseable output. Judge reliability is therefore load-bearing here in a way it is not for post-hoc explanation, since an extraction error propagates directly into the feature matrix the classifier is trained on. Related evidence from tabular settings shows LLMs favour simple, low-order feature operations~\citep{kuken2024simple}, which is consistent with the concrete, directly checkable predicates our generator is instructed to produce.

%
%

\section{Methods}
\label{sec_methods}

\subsection{Framework overview}

Our framework consists of three components: (1) a feature generator LLM that proposes natural language feature definitions, (2) a feature evaluator LLM that extracts binary feature values from text samples, and (3) a linear classifier trained on the resulting feature vectors. We use Claude Haiku 4.5 as both generator and evaluator. The feature set evolves over multiple generations using classification performance and feature importance as feedback signals.

Given a classification task with training samples $\{(x_i, y_i)\}_{i=1}^{N}$ and label set $\mathcal{Y}$, the system maintains a rubric $\mathbf{r} = \{r_1, r_2, \ldots, r_m\}$ of $m$ natural language feature definitions. Each feature $r_j$ is a short description of a binary attribute (e.g., ``Does the text contain explicit criticism of a product or service?''). For each sample $x_i$, the evaluator LLM produces a binary feature vector $\mathbf{f}_i \in \{0, 1\}^m$ indicating whether each feature is present. A logistic regression classifier is then trained on the feature matrix $\mathbf{F} \in \{0, 1\}^{N \times m}$ to predict labels $\mathbf{y}$. The input $x_i$ may be natural language text or structured data serialized to text (e.g., tabular features rendered as key-value descriptions).

\subsection{Feature evolution}

The evolutionary process proceeds over $T$ iterations. At each iteration $t$:

\begin{enumerate}
    \item \textbf{Evaluate}: Extract features from all training samples using the current rubric $\mathbf{r}^{(t)}$ and evaluate classification performance via $k$-fold cross-validation.
    \item \textbf{Analyze}: Compute feature importance scores from the classifier coefficients, per-class activation rates for each feature, and feature ablation scores (accuracy change when each feature is removed). Identify misclassified samples and confusion patterns.
    \item \textbf{Evolve}: Present the LLM with the current rubric, ablation scores (marking features as essential, useful, or unnecessary), per-class activation rates, and classification errors. The generator retains essential features and proposes replacements for unnecessary ones.
\end{enumerate}

\begin{figure}[t]
\begin{tcolorbox}[title=\textbf{Algorithm 1}: Evolutionary Feature Discovery, enhanced, colback=white, colframe=black!70, fonttitle=\small\bfseries, boxrule=0.5pt, top=2pt, bottom=2pt]
\small
\textbf{Input:} Training data $\mathcal{D}$, label set $\mathcal{Y}$, iterations $T$, retained top-$k$\\
\textbf{Output:} Optimised rubric $\mathbf{r}^*$
\begin{enumerate}[leftmargin=*, itemsep=1pt, topsep=2pt]
    \item $\mathbf{r}^* \gets \textsc{LLM-Generate}(\mathcal{Y}, \text{examples from } \mathcal{D})$
    \item $(s^*, \mathbf{w}, \mathcal{E}, \mathbf{a}, \mathbf{b}) \gets \textsc{Evaluate}(\mathbf{r}^*, \mathcal{D})$ \quad // CV score, coefficients, errors, activations, ablations
    \item $\mathcal{R} \gets \emptyset$ \quad // rejected candidates
    \item \textbf{for} $t = 1$ \textbf{to} $T$ \textbf{do}
    \begin{enumerate}[leftmargin=*, itemsep=0pt]
        \item[] $\mathcal{K} \gets$ top-$k$ features of $\mathbf{r}^*$ by $|\mathbf{w}|$ \quad // retained verbatim
        \item[] $\mathbf{r}' \gets \textsc{LLM-Generate}(\mathcal{Y}, \mathcal{K}, \mathbf{w}, \mathcal{E}, \mathbf{a}, \mathbf{b}, \mathcal{R})$
        \item[] $(s', \mathbf{w}', \mathcal{E}', \mathbf{a}', \mathbf{b}') \gets \textsc{Evaluate}(\mathbf{r}', \mathcal{D})$
        \item[] \textbf{if} $s' > s^*$ \textbf{then} $s^* \gets s'$; $\mathbf{r}^* \gets \mathbf{r}'$; $(\mathbf{w}, \mathcal{E}, \mathbf{a}, \mathbf{b}) \gets (\mathbf{w}', \mathcal{E}', \mathbf{a}', \mathbf{b}')$; $\mathcal{R} \gets \emptyset$
        \item[] \textbf{else} $\mathcal{R} \gets \mathcal{R} \cup \{\mathbf{r}'\}$
    \end{enumerate}
    \item \textbf{return} $\mathbf{r}^*$
\end{enumerate}
\end{tcolorbox}
\end{figure}

The generator is instructed to retain the top-$k$ features by importance, propose replacements for low-importance features, and generate new features that address observed error patterns. Previously rejected rubrics are fed back to the generator to steer it away from repeating unproductive candidates, and are cleared whenever a candidate improves the score. The full procedure is summarised in Algorithm~1, where \textsc{Evaluate} denotes feature extraction, $k$-fold cross-validation, and the analysis of step~2 (coefficients $\mathbf{w}$, errors $\mathcal{E}$, activations $\mathbf{a}$, ablations $\mathbf{b}$).

\subsection{Feature extraction and classification}

For each sample $x_i$ and rubric $\mathbf{r}$, the evaluator LLM is prompted to output a binary vector indicating the presence or absence of each feature. The prompt includes the full rubric and the text sample, and the LLM returns structured output in a parseable format (the full prompt template is provided in Appendix~\ref{app:prompts}). Feature extraction is parallelized across samples for efficiency.

We use logistic regression as the downstream classifier, chosen for its transparency: the learned coefficients directly indicate the contribution of each feature to each class prediction.\footnote{We use scikit-learn's \texttt{LogisticRegression} with $\ell_2$ regularisation ($C=0.1$), the \texttt{lbfgs} solver, multinomial loss for multiclass tasks, and \texttt{max\_iter}$=1000$. Features are binary, so no standardisation is applied.} The classifier is trained with $\ell_2$ regularisation and evaluated using stratified $k$-fold cross-validation during the evolution process, with a held-out test set for final evaluation.

\subsection{Baselines}

Figure~\ref{fig:comparison} illustrates the three LLM-based approaches we compare, alongside a TF-IDF baseline:

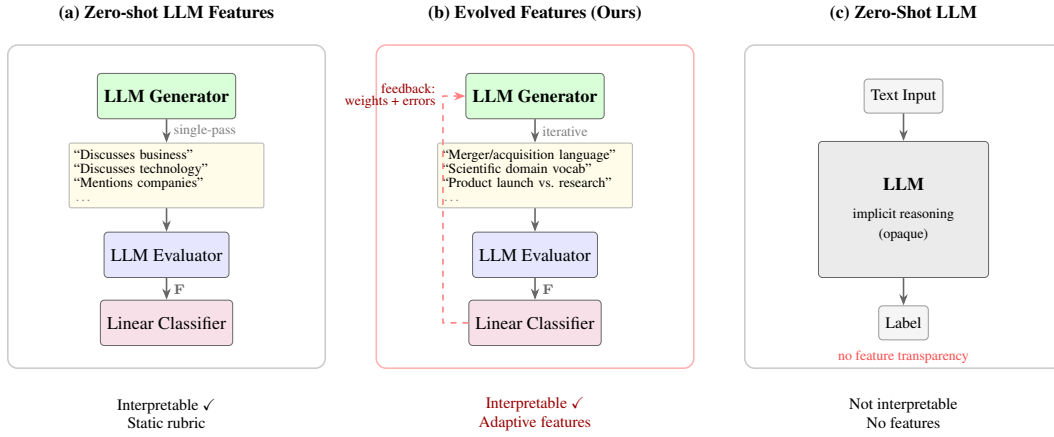
\begin{figure*}[t]
\centering
\resizebox{\textwidth}{!}{%
\begin{tikzpicture}[
    node distance=0.8cm and 0.6cm,
    box/.style={rectangle, draw=black!60, fill=#1, rounded corners=2pt,
                minimum height=0.8cm, minimum width=2.2cm, align=center, font=\small},
    box/.default={white},
    data/.style={rectangle, draw=black!40, fill=gray!8, rounded corners=2pt,
                 minimum height=0.6cm, align=center, font=\footnotesize},
    arr/.style={-{Stealth[length=2mm]}, thick, black!60},
    paneltitle/.style={font=\small\bfseries, anchor=south},
    feat/.style={rectangle, draw=black!30, fill=yellow!10, rounded corners=1pt,
                 font=\scriptsize, align=left, text width=3.2cm},
]

\begin{scope}[shift={(0,0)}]
  \node[paneltitle] at (2.2, 3.8) {(a) Zero-shot LLM Features};
  \draw[black!20, thick, rounded corners=5pt] (-0.6, -2.2) rectangle (5.0, 3.5);

  \node[box=green!15] (ha) at (2.2, 2.6) {\textbf{LLM Generator}};
  \node[feat] (hf) at (2.2, 1.2) {%
    {\scriptsize``Discusses business''}\\[-1pt]
    {\scriptsize``Discusses technology''}\\[-1pt]
    {\scriptsize``Mentions companies''}\\[-1pt]
    {\scriptsize\color{gray}\ldots}%
  };
  \node[box=blue!10] (he) at (2.2, -0.2) {LLM Evaluator};
  \node[box=purple!12] (hc) at (2.2, -1.4) {Linear Classifier};

  \draw[arr] (ha) -- (hf) node[midway, right, font=\scriptsize, text=black!50] {single-pass};
  \draw[arr] (hf) -- (he);
  \draw[arr] (he) -- (hc) node[midway, right, font=\scriptsize] {$\mathbf{F}$};
\end{scope}

\begin{scope}[shift={(6.5,0)}]
  \node[paneltitle] at (2.2, 3.8) {(b) Evolved Features (Ours)};
  \draw[red!30, thick, rounded corners=5pt] (-0.6, -2.2) rectangle (5.0, 3.5);

  \node[box=green!15] (ea) at (2.2, 2.6) {\textbf{LLM Generator}};
  \node[feat] (ef) at (2.2, 1.2) {%
    {\scriptsize``Merger/acquisition language''}\\[-1pt]
    {\scriptsize``Scientific domain vocab''}\\[-1pt]
    {\scriptsize``Product launch vs.\ research''}\\[-1pt]
    {\scriptsize\color{gray}\ldots}%
  };
  \node[box=blue!10] (ee) at (2.2, -0.2) {LLM Evaluator};
  \node[box=purple!12] (ec) at (2.2, -1.4) {Linear Classifier};

  \draw[arr] (ea) -- (ef) node[midway, right, font=\scriptsize, text=black!50] {iterative};
  \draw[arr] (ef) -- (ee);
  \draw[arr] (ee) -- (ec) node[midway, right, font=\scriptsize] {$\mathbf{F}$};

  \draw[-{Stealth[length=2mm]}, thick, red!50, dashed]
    (ec.west) -- ++(-0.45,0) |- (ea.west)
    node[pos=0.5, left, font=\scriptsize, text=red!60!black, align=right] {feedback:\\[-2pt]\scriptsize weights + errors};
\end{scope}

\begin{scope}[shift={(13,0)}]
  \node[paneltitle] at (2.2, 3.8) {(c) Zero-Shot LLM};
  \draw[black!20, thick, rounded corners=5pt] (-0.6, -2.2) rectangle (5.0, 3.5);

  \node[data] (ci) at (2.2, 2.6) {Text Input};
  \node[box=black!8, minimum height=2.4cm, minimum width=3.0cm] (cb) at (2.2, 0.6) {%
    \textbf{LLM}\\[4pt]
    {\scriptsize implicit reasoning}\\
    {\scriptsize (opaque)}%
  };
  \node[data] (co) at (2.2, -1.4) {Label};

  \draw[arr] (ci) -- (cb);
  \draw[arr] (cb) -- (co);

  \node[font=\scriptsize, text=red!70, align=center] at (2.2, -2.0) {no feature transparency};
\end{scope}

\node[font=\footnotesize, align=center] at (2.2, -3.0) {Interpretable \checkmark\\Static rubric};
\node[font=\footnotesize, align=center, text=red!60!black] at (8.7, -3.0) {Interpretable \checkmark\\Adaptive features};
\node[font=\footnotesize, align=center] at (15.2, -3.0) {Not interpretable\\No features};

\end{tikzpicture}%
}
\caption{Comparison of three LLM classification approaches. (a)~An LLM generates a static rubric of feature definitions in a single pass, evaluated by an LLM. (b)~Our framework uses an LLM to iteratively generate and refine features via evolutionary feedback from classifier performance. (c)~A zero-shot LLM classifies directly without explicit features. Only (a) and (b) produce transparent, auditable classifiers; only (b) adapts features to the data.}
\label{fig:comparison}
\end{figure*}

\begin{itemize}
    \item \textbf{TF-IDF + logistic regression}: A bag-of-words baseline using term frequency-inverse document frequency features with $\ell_2$-regularised logistic regression. Fully transparent (individual word weights are inspectable) but limited to surface-level lexical patterns.
    \item \textbf{Zero-shot features} (Figure~\ref{fig:comparison}a): The LLM generates a rubric of $m$ binary features in a single pass (iteration 0 of the evolutionary process), given the task description, label set, and a sample of labelled examples. No iterative refinement or performance feedback is used.
    \item \textbf{Evolved features} (proposed approach; Figure~\ref{fig:comparison}b): Features refined through the proposed evolutionary framework over $T$ iterations, starting from the zero-shot rubric.
    \item \textbf{Zero-shot LLM classifier} (Figure~\ref{fig:comparison}c): The LLM classifies each sample directly with a single prompt, using no learned features. This black-box baseline offers no transparency but is expected to perform well given its flexibility.
\end{itemize}

%
%

\section{Experimental setup and results}
\label{sec_results}

\subsection{Datasets}

We evaluate our evolutionary feature discovery framework across three classification benchmarks spanning different task types, difficulty levels, and domains:

\begin{itemize}
    \item \textbf{AG News}~\citep{zhang2015character}: 4-class news topic classification (World, Sports, Business, Sci/Tech). Well-separated categories with distinctive vocabulary.
    \item \textbf{GoEmotions}~\citep{demszky2020goemotions}: 7-class emotion classification (online comments mapped from 28 fine-grained labels to Ekman emotions). Complex taxonomy with arbitrary label boundaries.
    \item \textbf{HELOC}~\citep{fico2018heloc}: Binary credit risk classification (at-risk vs.\ not-at-risk for default on a Home Equity Line of Credit). Tabular credit report features serialized to natural language text. This dataset represents a regulated domain where interpretability is legally required and where LLM domain competence cannot be assumed.
\end{itemize}

\subsection{Experimental configuration}

The evolutionary search uses $n=500$ samples per dataset with an 80/20 stratified train/test split, $T=15$ iterations, 5-fold cross-validation for rubric selection, and rubrics of $m=10$ binary features. We use Claude Haiku 4.5 as both the feature generator and evaluator, with $\ell_2$-regularised logistic regression ($C=0.1$) as the downstream classifier.

At each iteration, the LLM receives three forms of feedback: (1) classification errors showing confused class pairs, (2) per-class activation rates identifying non-discriminative features, and (3) ablation scores marking features essential, useful, or unnecessary, with instructions to keep the essential features and replace the rest.

To evaluate generalisation, we assess both the initial and final evolved rubrics on 3 independent stratified splits and report mean $\pm$ std. All baselines use identical data splits for fair comparison.

\subsection{Main results}

\begin{table}[t]
  \centering
  \caption{Test accuracy (\%) across 3 evaluation splits (mean $\pm$ std). Best per dataset in \textbf{bold}.}
  \label{tab:main_results}
  \begin{tabularx}{\linewidth}{@{} l *{4}{>{\centering\arraybackslash}X} @{}}
  \toprule
  Dataset & TF-IDF & Zero-shot features & Evolved features & Zero-shot LLM \\
  \midrule
  AG News     & 80.0{\scriptsize$\pm$4.1} & 82.7{\scriptsize$\pm$0.5} & 81.3{\scriptsize$\pm$2.6} & \textbf{89.7}{\scriptsize$\pm$1.2} \\
  GoEmotions  & 31.3{\scriptsize$\pm$8.2} & 34.7{\scriptsize$\pm$6.2} & \textbf{41.0}{\scriptsize$\pm$1.6} & 38.7{\scriptsize$\pm$3.7} \\
  HELOC       & 62.3{\scriptsize$\pm$6.0} & 65.3{\scriptsize$\pm$5.7} & \textbf{69.0}{\scriptsize$\pm$2.2} & 50.7{\scriptsize$\pm$1.7} \\
  \midrule
  Average & 57.9 & 60.9 & \textbf{63.8} & 59.7 \\
  \bottomrule
  \end{tabularx}
\end{table}

Table~\ref{tab:main_results} presents the main results. The methods trade transparency against accuracy: TF-IDF offers full word-level interpretability but limited semantic understanding, while the zero-shot LLM classifier achieves the highest accuracy only on the one task with self-explanatory categories. Our evolved features achieve the best average accuracy across all three benchmarks while staying human-readable.

Evolved features improve over single-shot rubrics on all tasks except AG News (+2.9 pp average), with the largest gains on GoEmotions (+6.3 pp) and HELOC (+3.7 pp). On two of three tasks, evolved features also outperform the zero-shot LLM classifier while providing full interpretability. When label boundaries are non-obvious or the LLM lacks domain reasoning, the search finds discriminative patterns its implicit reasoning misses.

The HELOC result is particularly striking: the zero-shot LLM reaches only 50.7\% accuracy, barely above chance, because it defaults to a strong ``at\_risk'' bias (77--85\% of predictions across seeds) when presented with structured credit data it cannot reliably reason about. This lopsided prediction pattern yields a high ``at\_risk'' recall (0.80--0.84) but poor ``not\_at\_risk'' recall (0.14--0.26), so the two classes are far from balanced. In contrast, evolved features achieve 69.0\% with balanced per-class predictions (per-class F1 of 0.64--0.71 for both classes across seeds), demonstrating that error-informed feature discovery produces functional classifiers even in domains where the LLM's zero-shot reasoning does not.

The GoEmotions result is also notable: evolved features (41.0\%) outperform both the zero-shot LLM (38.7\%) and TF-IDF (31.3\%), suggesting that the 7-class Ekman emotion mapping creates a taxonomy that benefits from targeted feature discovery.

On AG News, the zero-shot LLM dominates (89.7\%) because the four categories are well-separated and self-explanatory. Evolution does not improve on the initial rubric (81.3\% vs 82.7\%): refinement adds noise rather than signal when a single-shot rubric already captures the distinctions.

Across all tasks, the results reveal a clear pattern: evolutionary refinement is most valuable when the classification task requires features that cannot be inferred from label names alone, either because the taxonomy is arbitrary (GoEmotions) or because the domain requires specialised numerical reasoning that the LLM cannot perform zero-shot (HELOC). Evolved features also show substantially lower variance than zero-shot features on GoEmotions ($\pm$1.6 vs $\pm$6.2), suggesting that the evolutionary search produces more robust feature sets even when the initial rubric quality varies across data splits.

%
%

\section{Analysis}
\label{sec_analysis}

\subsection{Convergence analysis}

\begin{figure}[t]
\centering
\includegraphics[width=0.75\columnwidth]{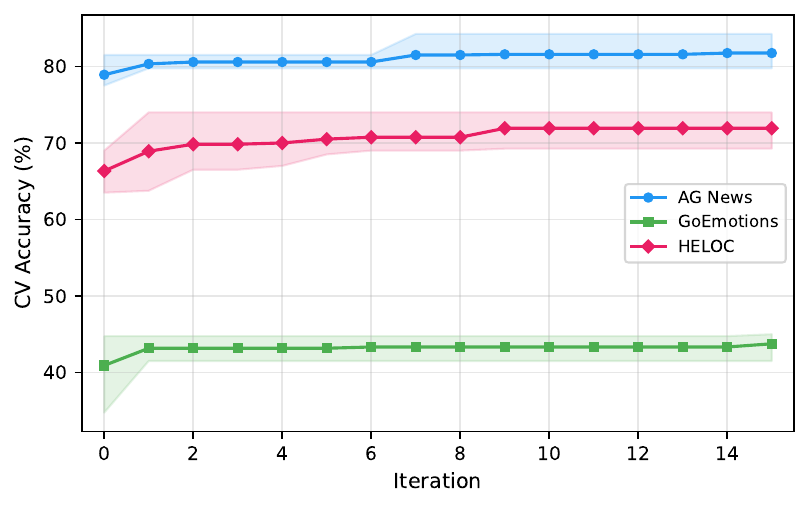}
\caption{Cross-validation accuracy over evolutionary iterations. Solid lines show best-accepted CV score; shaded regions show the range across 3 seeds. Evolution makes decisive improvements early in the search on tasks where the initial rubric is suboptimal.}
\label{fig:convergence}
\end{figure}

Figure~\ref{fig:convergence} shows the search trajectory for each dataset. On GoEmotions, the first accepted rubric yields substantial improvement (+10 to +12 pp), after which the search plateaus. HELOC follows a similar pattern: CV rises from a range of 0.65 to 0.69 into a range of 0.69 to 0.74 within the first few accepted iterations, then plateaus. This pattern of rapid initial gain followed by diminishing returns suggests that the primary value of evolution lies in the first feedback-informed refinements: the LLM observes classification errors from the initial rubric and proposes a substantially better feature set. Subsequent iterations rarely improve further, as the bar for acceptance rises quickly.
AG News shows no meaningful improvement, confirming that well-separated categories are adequately captured by a single-shot rubric without iterative refinement.

Based on these convergence patterns, we recommend $T=5$ to $10$ iterations as sufficient for most tasks. Across all three datasets, the vast majority of gains are realised within the first 5 accepted rubrics, though occasional late improvements occur (e.g., one HELOC seed accepted an improvement at iteration 9). Running beyond 10 iterations incurs additional LLM cost without measurable benefit.

\subsection{Cost-accuracy trade-off}

\begin{table}[b]
  \centering
  \caption{LLM calls required for training (feature discovery) and inference (per sample). Evolved features assume $T=10$ iterations with $n=500$ training samples.}
  \label{tab:cost}
  \begin{tabularx}{\linewidth}{@{} X >{\raggedleft\arraybackslash}p{3.4cm} >{\raggedleft\arraybackslash}p{3.0cm} @{}}
  \toprule
  Method & Classifier training calls & LLM inference calls \\
  \midrule
  TF-IDF + logistic regression & 0 & 0 \\
  Zero-shot features & 501 & 1 \\
  Evolved features (ours) & 5{,}010 & 1 \\
  Zero-shot LLM classifier & 0 & 1 \\
  \bottomrule
  \end{tabularx}
\end{table}

Table~\ref{tab:cost} shows the evolved features approach requires approximately 5{,}000 LLM calls during training ($T$ iterations $\times$ $n$ extraction calls per iteration, plus $T$ generation calls). However, at inference time, all LLM-based feature methods require exactly one call per sample, the same as the zero-shot LLM classifier. Since inference dominates cost after deployment (training is amortised over all future predictions), the evolved and zero-shot feature approaches have identical operational cost. The extra ${\sim}5{,}000$ training calls buy +2.9 pp over zero-shot features at no added inference cost.

\subsection{Feature evolution}

We analyse the evolved rubrics to understand what evolution discovers (full visualisation in Figure~\ref{fig:survival}, Appendix~\ref{app:feature_evolution}). Across all three datasets, evolution aggressively replaces the initial rubric: AG News retains only 1 of 10 initial features, GoEmotions retains 3, and HELOC retains 2. The replacements follow task-specific strategies. On text tasks, evolution replaces generic, broadly-firing features with specialised discriminators targeting specific confusion pairs, for example, replacing ``Presence of exclamation marks'' (which fires across multiple emotion classes) with ``Presence of concrete threat or aggressive action framing'' (targeting anger specifically). On HELOC, evolution discovers numerical threshold features (``Nr Trades Insolvent For Over 60 Days $\geq$ 1'') that are effectively decision stumps expressed in natural language, binary splits that a domain expert can immediately validate against regulatory guidelines.

\subsection{Interpretability}

A key advantage of the framework is end-to-end transparency. For any prediction, a practitioner can inspect: (1) the 10 feature definitions in natural language, (2) the binary feature values extracted for the input, and (3) the logistic regression coefficients linking features to classes.

\begin{figure}[t]
  \centering
  \begin{subfigure}{0.49\linewidth}
    \centering
    \includegraphics[width=\linewidth]{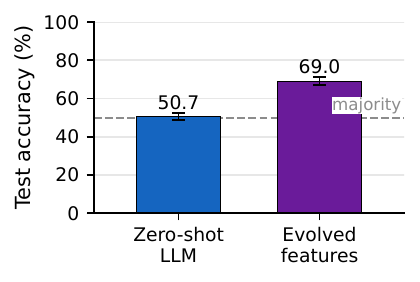}
    \caption{Aggregate view}
    \label{fig:heloc_audit_a}
  \end{subfigure}
  \hfill
  \begin{subfigure}{0.49\linewidth}
    \centering
    \includegraphics[width=\linewidth]{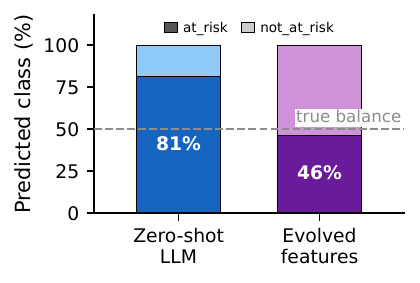}
    \caption{Per-class audit}
    \label{fig:heloc_audit_b}
  \end{subfigure}
  \caption{Aggregate accuracy hides a per-class failure on HELOC. (a)~Both methods sit near or above the majority baseline, so neither looks broken. (b)~On the balanced 50/50 test set, the zero-shot LLM predicts \texttt{at\_risk} for 81\% of applicants (a systematic single-class bias), whereas evolved features stay balanced. The failure surfaces only under per-class auditing.}
  \label{fig:heloc_audit}
\end{figure}

This distinction matters for evaluation practice: Figure~\ref{fig:heloc_audit} shows that the zero-shot LLM assigns \texttt{at\_risk} to 81\% of a balanced test set, a systematic bias that its near-chance aggregate accuracy conceals but the interpretable feature layer exposes.

\begin{table}[t]
  \centering
  \caption{Evolved rubric for HELOC, the task on which the zero-shot LLM performs at chance with a strong single-class bias. Evolution discovers numerical thresholds that act as decision stumps expressed in natural language, each checkable against a credit report.}
  \label{tab:evolved_rubric_heloc}
  \small
  \begin{tabularx}{\linewidth}{@{} c X >{\raggedleft\arraybackslash}p{1.0cm} @{}}
    \toprule
    \# & Feature definition & $|\bar{w}|$ \\
    \midrule
    1 & Net Fraction Of Revolving Burden $>$ 40 (high revolving debt) & 0.652 \\
    2 & Number Of Satisfactory Trades $<$ 15 (few satisfactory trades) & 0.539 \\
    3 & Months Since Last Illegal Trade between 0 and 12 (recent illegal-trade activity) & 0.505 \\
    4 & Percentage Of Legal Trades $<$ 90 (low legal-trade share) & 0.399 \\
    5 & Months Since First Trade $<$ 100 and Nr Trades Initiated In Last Year $\geq$ 3 (short history with recent activity) & 0.374 \\
    6 & Nr Banks With High Ratio $\geq$ 2 (multiple lenders stressed) & 0.314 \\
    7 & Nr Trades Insolvent For Over 60 Days $\geq$ 1 (moderate delinquency history) & 0.285 \\
    8 & Nr Inquiries In Last 6 Months $\geq$ 3 and Nr Total Trades $\leq$ 25 (excessive inquiries relative to portfolio) & 0.131 \\
    9 & Nr Total Trades $\leq$ 5 and Nr Trades Initiated In Last Year $\geq$ 2 (minimal portfolio with high recent activity) & 0.088 \\
    10 & Net Fraction Of Installment Burden $>$ 85 (extreme installment burden) & 0.050 \\
    \bottomrule
  \end{tabularx}
\end{table}

This transparency is especially valuable in regulated domains. Table~\ref{tab:evolved_rubric_heloc} gives the full evolved HELOC rubric, the decision logic behind every prediction on the task where the zero-shot LLM falls to chance. A credit decision reads as: ``Applicant classified as at-risk because: Net Fraction Of Revolving Burden $>$ 40 (high revolving debt), Nr Trades Insolvent For Over 60 Days $\geq$ 1 (moderate delinquency history).'' Such explanations are structured to support the adverse action notices required by the Equal Credit Opportunity Act, though establishing their legal validity would require fairness, proxy-discrimination, and domain-expert auditing beyond our scope. Every row is a threshold a credit officer can check against a report and dispute, which is what makes the failure of the zero-shot baseline visible in the first place. Rubrics for the two text tasks appear in Appendix~\ref{app:rubrics}.

%
%

\section{Conclusion}
\label{sec_conclusion}
\label{sec:broader_impact}

We presented an evolutionary framework that uses an LLM to iteratively discover natural-language features for interpretable classification, producing classifiers whose decision logic is human-readable and auditable. Across three benchmarks, evolved features improve over single-shot rubrics by +2.9 pp on average and outperform zero-shot LLM classification on two of three tasks. On the HELOC credit-risk task, where the zero-shot LLM performs at chance with a strong single-class bias, they remain balanced and auditable. Evolution helps most when label boundaries cannot be inferred from category names alone, or when the LLM lacks domain reasoning.

Evolution adds little when a single-shot rubric is already near-optimal or the LLM's general knowledge suffices for direct classification. The binary representation also bottlenecks fine-grained tasks, and our analysis uses a single LLM (Claude Haiku 4.5), leaving cross-model generalisation to future work. Other directions include broader evaluation on financial and medical data, feature-pool architectures that decouple generation from L1-regularised selection, and targeted mutation of individual features rather than whole-rubric regeneration.

Two implications for evaluation practice follow. First, aggregate accuracy can certify a classifier that is near chance for the wrong reason: the zero-shot LLM's 50.7\% on HELOC reads as task difficulty, yet the model is heavily single-class-biased, and only per-class auditing separates the two cases. Per-class metrics should thus precede deployment, and a transparent representation makes that audit routine. Second, feature values come from an LLM judge, so a miscalibrated extractor can corrupt otherwise transparent logic; extraction accuracy needs validating separately from end-task accuracy.


\FloatBarrier

\bibliographystyle{plainnat}
\bibliography{references}

\clearpage
\appendix
%
%

\section{Prompt templates}
\label{app:prompts}

\subsection{Feature extraction prompt}

\begin{tcolorbox}[title=\textbf{Feature Extraction}, enhanced, breakable, colback=gray!5, colframe=gray!50!black, fonttitle=\bfseries, boxrule=0.4pt]
\small
Extract the provided binary features from the text using the rubrics below.
Output a list of exactly \{num\_features\} values (0s and 1s, one per feature, in order) inside \texttt{<features></features>} tags.

\textbf{Rubric:}\\
1. \{feature\_1\}\\
2. \{feature\_2\}\\
\ldots

\textbf{Text:}\\
\{text\}

\textbf{Format:}\\
\texttt{<features>}\\
0,0,\ldots\\
\texttt{</features>}
\end{tcolorbox}

\subsection{Feature evolution prompt}

The evolution prompt includes the task description, label set, a sample of labelled training examples, the current top-performing features with their classifier coefficients, error analysis from the previous iteration (confusion pairs and sample misclassifications), per-class feature activation rates, feature ablation scores marking each feature as essential, useful, or unnecessary, and any recently rejected rubrics. A simplified version is shown below; the full prompt is available in the code repository.

\begin{tcolorbox}[title=\textbf{Feature Evolution} (simplified), enhanced, breakable, colback=gray!5, colframe=gray!50!black, fonttitle=\bfseries, boxrule=0.4pt]
\small
Your task is to generate a rubric with 10 binary features which will be used as ML model features to determine a classification. Keep the top performing existing features and replace/iterate on the others to find new complementary features that will improve performance. We want generalisable features that transfer to unseen examples, not only training instances.

Each feature MUST be concrete and directly extractable: check for presence of specific words, phrases, named entities, or syntactic patterns. Do NOT use features that require counting, computing ratios, or subjective judgment. Think step-by-step about the evidence before generating your final rubric.

\textbf{Problem Domain:} \{task\_description\}\\
\textbf{Possible Labels:} \{labels\}

\textbf{Example input/output pairs:}\\
\{labelled\_examples\}

\textbf{Top performing features:}\\
\{top\_features\_with\_coefficients\}

\textbf{Previous iteration results:}\\
CV accuracy: \{cv\_accuracy\}\\
Most common confusions: \{confusion\_pairs\}\\
Sample errors: \{misclassified\_examples\}\\
Feature importance: \{feature\_coefficients\}\\
Per-class activation rates: \{activation\_rates\}\\
Feature necessity (ablation): \{ablation\_scores\}\\
Replace features marked UNNECESSARY. Keep features marked ESSENTIAL unchanged.

\textbf{Previously rejected rubrics:}\\
\{rejected\_rubrics\}
\end{tcolorbox}

\section{Feature evolution analysis}
\label{app:feature_evolution}

\begin{figure*}[b]
\centering
\includegraphics[width=0.95\textwidth]{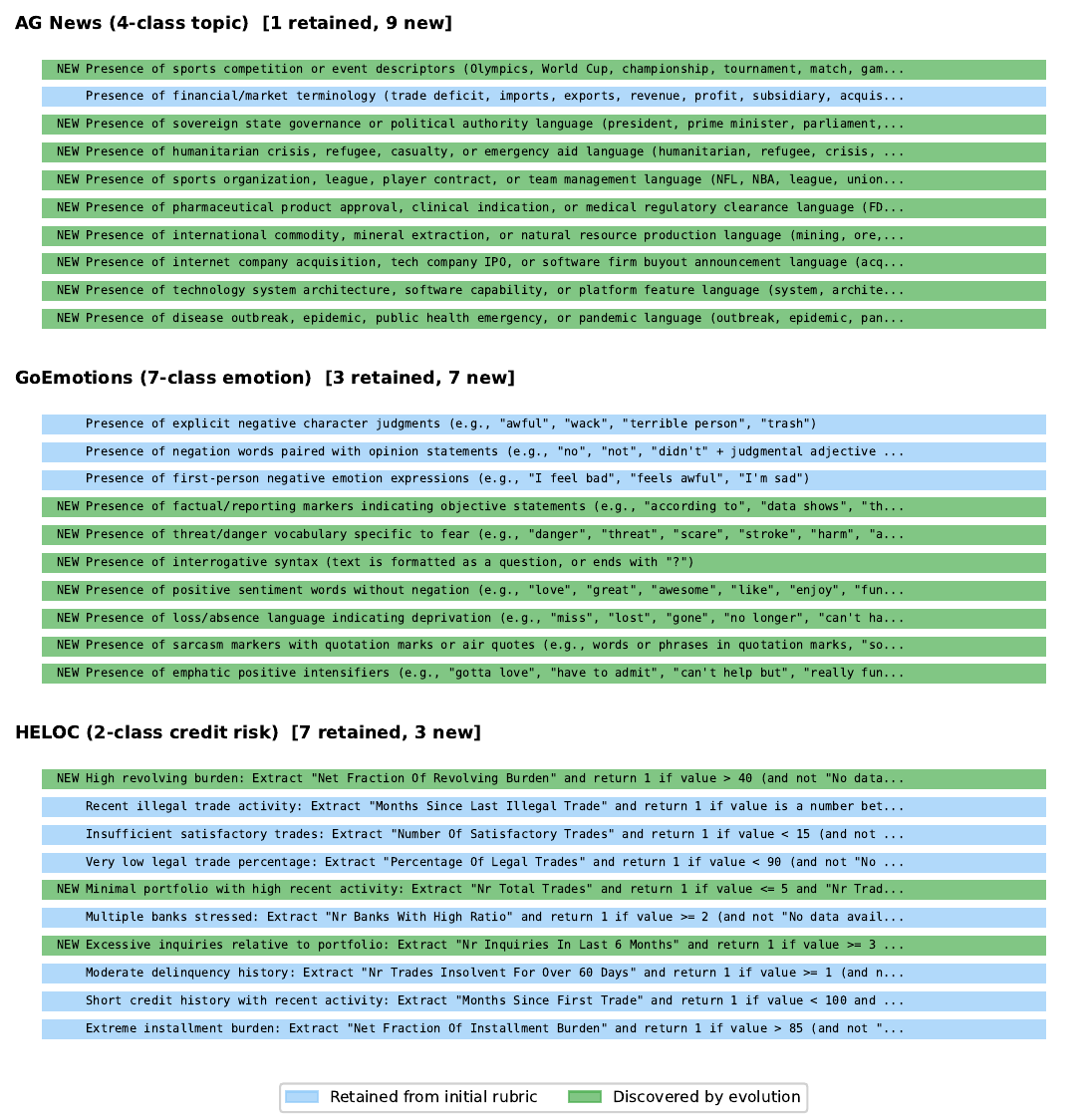}
\caption{Evolved rubrics for all three datasets, showing which features were retained from the initial rubric (blue) and which were newly discovered by evolution (green). Evolution replaces generic features with task-specific discriminators, with the degree of replacement reflecting task difficulty.}
\label{fig:survival}
\end{figure*}

Figure~\ref{fig:survival} shows the final evolved rubrics with features colour-coded by origin. The replacements reveal a consistent strategy: the LLM identifies which class pairs are most confused and proposes features that resolve those specific confusions. On the emotion task, generic indicators like ``exclamation marks'' fire across multiple classes and provide little discriminative signal. Evolution replaces them with targeted detectors: explicit sadness markers, sarcasm via negation patterns, rhetorical questions for surprise, and hostile second-person address for anger. Each is designed to separate a specific pair of emotions that the initial rubric conflated. The retained features (positive markers, anxiety words, moral disapproval) survive precisely because they already map cleanly onto individual classes.

On HELOC, the zero-shot LLM performs at chance with a strong single-class bias because it cannot perform reliable quantitative reasoning over serialized numerical profiles, whereas evolution uses classifier feedback to identify which thresholds discriminate between risk classes. The evolved features are compound conditions (e.g., ``Nr Inquiries In Last 6 Months $\geq$ 3 AND Nr Total Trades $\leq$ 25'') that capture interaction effects between credit variables.

\section{Evolved rubric examples}
\label{app:rubrics}

\begin{table*}[b]
  \centering
  \caption{Final evolved rubric for AG News (4-class topic classification) after 15 iterations. The initial rubric (iteration 0) was never improved upon. Features ordered by average absolute classifier coefficient.}
  \label{tab:evolved_rubric_agnews}
  \small
  \begin{tabularx}{\textwidth}{@{} c X >{\raggedleft\arraybackslash}p{1.2cm} @{}}
    \toprule
    \# & Feature definition & $|\bar{w}|$ \\
    \midrule
    1 & Contains sports-specific terminology (athlete names, team abbreviations, sport league names like NHL, NBA, or position/play descriptions) & 0.854 \\
    2 & Contains financial or market-specific terms (deficit, trade, stocks, revenue, earnings, merger, acquisition, deal) & 0.622 \\
    3 & Contains corporate or brand entity names (company names that are widely recognised business organisations) & 0.576 \\
    4 & Contains equipment, vehicle, or technology product names/models (gadgets, devices, software names, satellites, hardware) & 0.470 \\
    5 & Contains language of diplomatic conflict or legal action (accuse, sue, claim, testify, violation, treaty, embargo, sanction) & 0.467 \\
    6 & Contains scientific or technology terminology (algorithm, software, digital, innovation, research, quantum, genetic, climate, renewable) & 0.441 \\
    7 & Contains player/team performance statistics or game scores (numbers paired with athletic metrics like goals, assists, wins, points) & 0.339 \\
    8 & Mentions specific scientific studies, research findings, or experimental results with descriptive outcomes & 0.122 \\
    9 & Contains references to international locations, countries, or geopolitical entities & 0.117 \\
    10 & Contains entertainment industry references or Hollywood/media production terminology & 0.078 \\
    \bottomrule
  \end{tabularx}
\end{table*}

\begin{table*}[t]
  \centering
  \caption{Evolved rubric for GoEmotions (7-class emotion). Features target specific emotion categories rather than generic sentiment indicators.}
  \label{tab:evolved_rubric_goemotions}
  \small
  \begin{tabularx}{\textwidth}{@{} c X >{\raggedleft\arraybackslash}p{1.2cm} @{}}
    \toprule
    \# & Feature definition & $|\bar{w}|$ \\
    \midrule
    1 & Contains anxiety/worry-related words: nervous, worried, anxious, scared, terrified & 0.384 \\
    2 & Contains explicit sadness or loss markers: sad, disappointed, loss, heartbroken & 0.370 \\
    3 & Contains positive outcome or satisfaction markers: works, good, great, nice, love & 0.362 \\
    4 & Contains explicit moral disapproval or dismissal: freaky, weird, inappropriate, awful & 0.347 \\
    5 & Contains rhetorical question structure without interrogative intent & 0.221 \\
    6 & Contains unfamiliarity discovery markers: didn't know, never knew, how is, what is & 0.168 \\
    7 & Contains direct second-person hostile address: you're, you people, you all & 0.167 \\
    8 & Contains negation followed by positive words indicating sarcasm: not great, not nice & 0.154 \\
    9 & Contains casual dismissal or eye-roll language: yeah right, sure, whatever, ok & 0.115 \\
    10 & Contains regret or nostalgia markers: miss, used to, back when, wish I had & 0.108 \\
    \bottomrule
  \end{tabularx}
\end{table*}

\section{Detailed analysis}
\label{app:analysis}

\subsection{Feature-class coefficient heatmaps}

\begin{figure*}[b]
\centering
\includegraphics[width=0.85\textwidth]{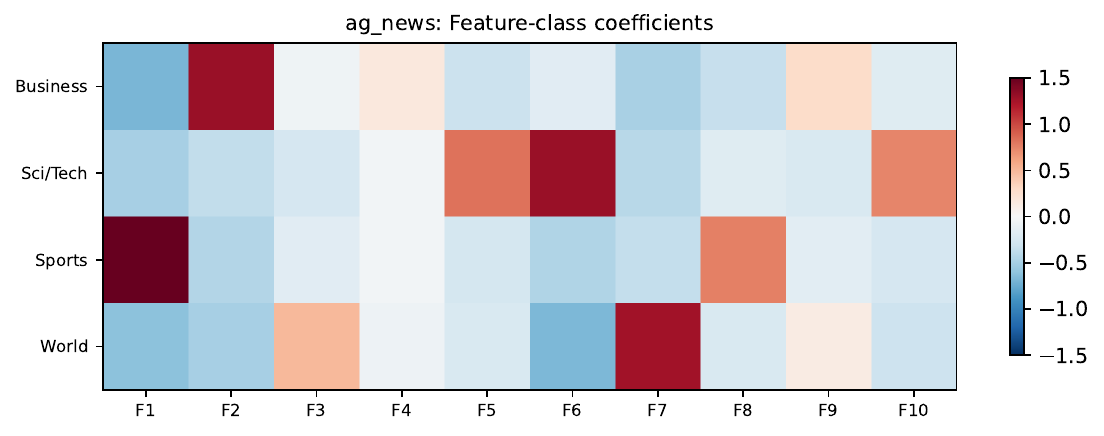}
\caption{AG News: feature-class coefficients.}
\label{fig:heatmap_ag}
\end{figure*}

\begin{figure*}[t]
\centering
\includegraphics[width=0.85\textwidth]{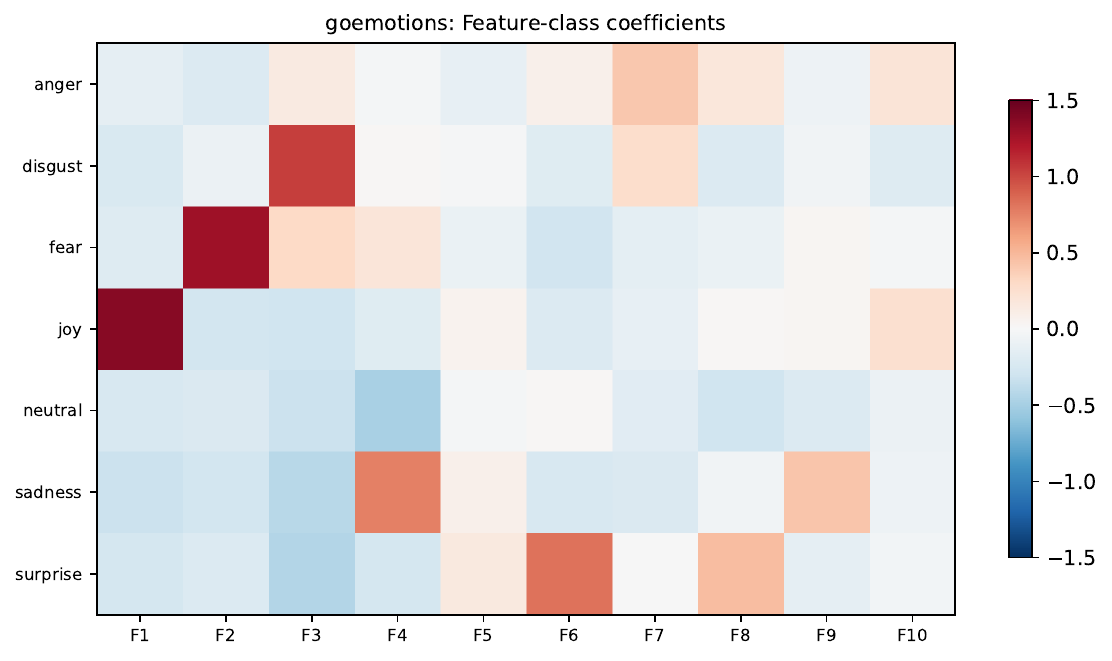}
\caption{GoEmotions: feature-class coefficients.}
\label{fig:heatmap_ge}
\end{figure*}

Figures~\ref{fig:heatmap_ag}--\ref{fig:heatmap_ge} show the logistic regression coefficients for each evolved feature (columns) and class (rows). Red indicates the feature increases the probability of that class; blue indicates it decreases it.

\subsection{Confusion matrix comparison}

\begin{figure*}[t]
\centering
\includegraphics[width=0.85\textwidth]{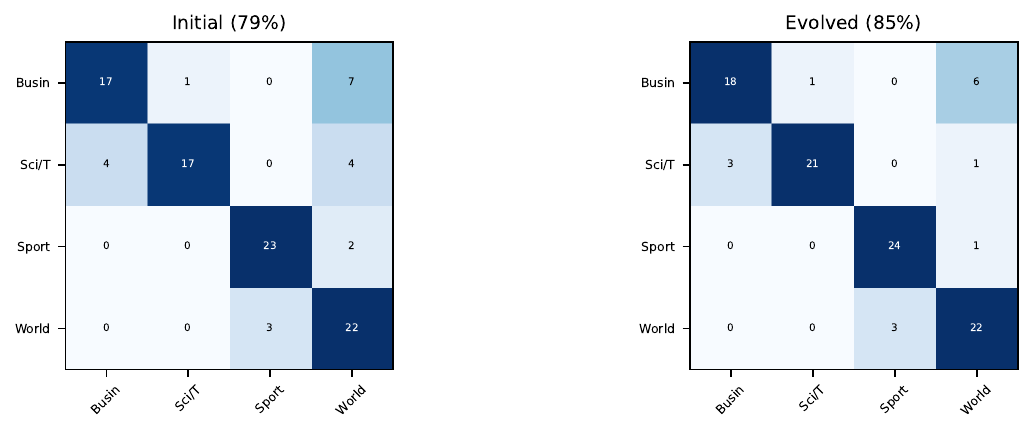}
\caption{AG News: confusion matrices (initial vs evolved).}
\label{fig:conf_ag}
\end{figure*}

\begin{figure*}[t]
\centering
\includegraphics[width=0.85\textwidth]{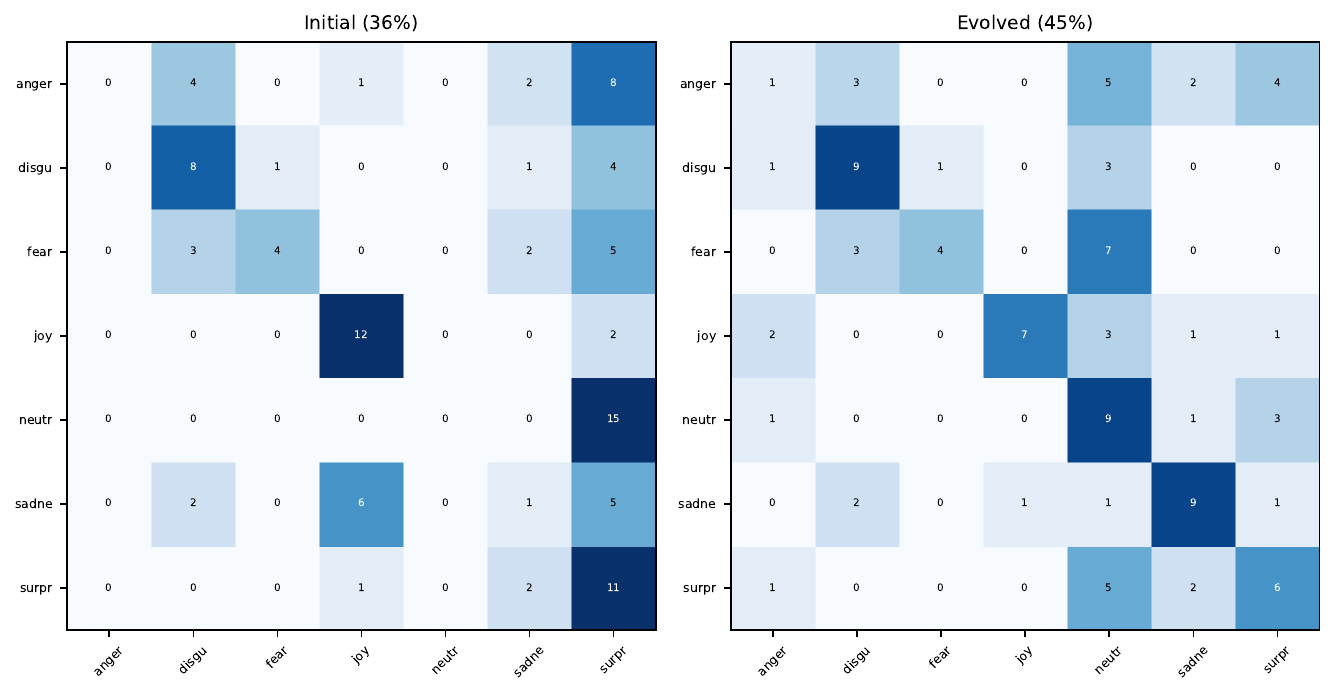}
\caption{GoEmotions: confusion matrices (initial vs evolved).}
\label{fig:conf_ge}
\end{figure*}

Figures~\ref{fig:conf_ag}--\ref{fig:conf_ge} compare the confusion matrices of the initial (zero-shot) rubric and the evolved rubric on held-out test data.

\subsection{Feature necessity ablation}

\begin{figure*}[t]
\centering
\includegraphics[width=0.85\textwidth]{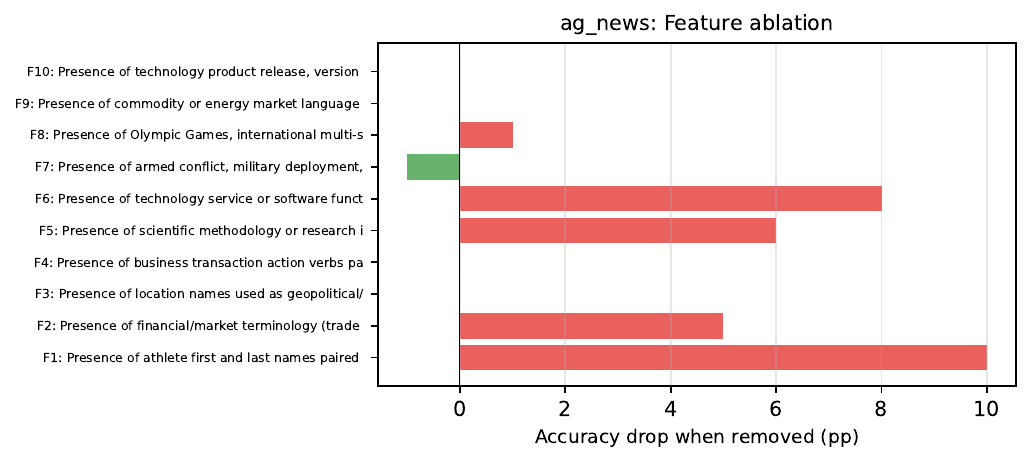}
\caption{AG News: feature ablation.}
\label{fig:abl_ag}
\end{figure*}

\begin{figure*}[t]
\centering
\includegraphics[width=0.85\textwidth]{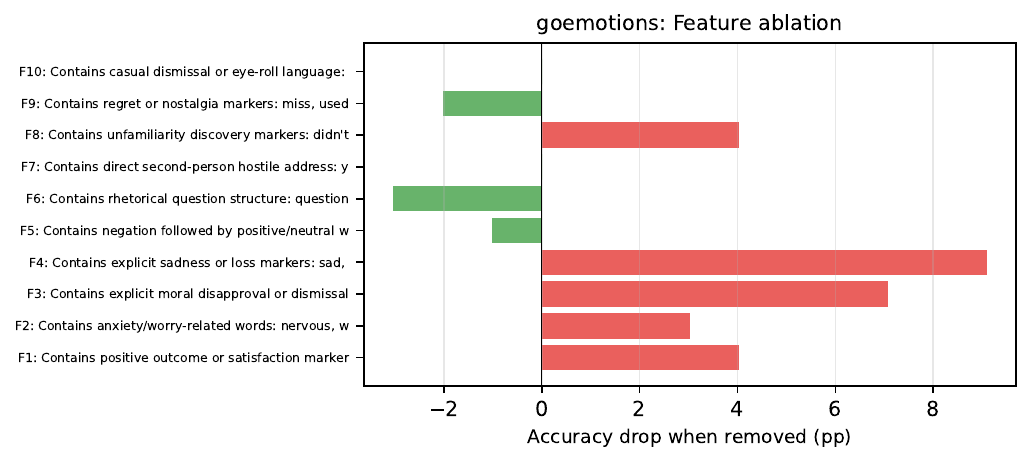}
\caption{GoEmotions: feature ablation.}
\label{fig:abl_ge}
\end{figure*}

Figures~\ref{fig:abl_ag}--\ref{fig:abl_ge} show the accuracy drop when each feature is removed from the evolved rubric. Red bars indicate features that contribute positively; green bars indicate features whose removal improves accuracy.

\clearpage
\section*{NeurIPS Paper Checklist}

\begin{enumerate}

\item {\bf Claims}
    \item[] Question: Do the main claims made in the abstract and introduction accurately reflect the paper's contributions and scope?
    \item[] Answer: \answerYes{} 
    \item[] Justification: The claims in the abstract and introduction (a +2.9 pp average gain over single-shot rubrics, outperforming the zero-shot LLM on two of three tasks, and the zero-shot LLM's chance-level single-class-biased performance on HELOC) are all substantiated by the results in Section~\ref{sec_results}.

\item {\bf Limitations}
    \item[] Question: Does the paper discuss the limitations of the work performed by the authors?
    \item[] Answer: \answerYes{} 
    \item[] Justification: Limitations are discussed in Section~\ref{sec_conclusion}: little benefit when a single-shot rubric is already near-optimal, the information bottleneck of binary features, and reliance on a single LLM.

\item {\bf Theory assumptions and proofs}
    \item[] Question: For each theoretical result, does the paper provide the full set of assumptions and a complete (and correct) proof?
    \item[] Answer: \answerNA{} 
    \item[] Justification: The paper presents an empirical framework and contains no formal theoretical results.

    \item {\bf Experimental result reproducibility}
    \item[] Question: Does the paper fully disclose all the information needed to reproduce the main experimental results of the paper to the extent that it affects the main claims and/or conclusions of the paper (regardless of whether the code and data are provided or not)?
    \item[] Answer: \answerYes{} 
    \item[] Justification: Section~\ref{sec_methods} and Section~\ref{sec_results} specify the algorithm, datasets, train/test splits, and all hyperparameters, and Appendix~\ref{app:prompts} provides the prompt templates needed to reproduce the results.

\item {\bf Open access to data and code}
    \item[] Question: Does the paper provide open access to the data and code, with sufficient instructions to faithfully reproduce the main experimental results, as described in supplemental material?
    \item[] Answer: \answerNo{} 
    \item[] Justification: We do not release source code. Reproducibility is instead supported by other means: all three datasets (AG News, GoEmotions, HELOC) are publicly available and cited, and we provide the full algorithm, train/test splits, all hyperparameters, and the prompt templates (Appendix~\ref{app:prompts}) needed to reproduce the results.

\item {\bf Experimental setting/details}
    \item[] Question: Does the paper specify all the training and test details (e.g., data splits, hyperparameters, how they were chosen, type of optimizer) necessary to understand the results?
    \item[] Answer: \answerYes{} 
    \item[] Justification: The experimental configuration in Section~\ref{sec_results} reports sample size, splits, number of iterations, cross-validation folds, feature count, classifier regularisation, and the LLM used.

\item {\bf Experiment statistical significance}
    \item[] Question: Does the paper report error bars suitably and correctly defined or other appropriate information about the statistical significance of the experiments?
    \item[] Answer: \answerYes{} 
    \item[] Justification: Tables in Section~\ref{sec_results} report mean $\pm$ standard deviation across three independent stratified splits.

\item {\bf Experiments compute resources}
    \item[] Question: For each experiment, does the paper provide sufficient information on the computer resources (type of compute workers, memory, time of execution) needed to reproduce the experiments?
    \item[] Answer: \answerYes{} 
    \item[] Justification: Cost is dominated by LLM API calls, reported per training run and per inference in Table~\ref{tab:cost}; the logistic-regression classifier is trained on 500 samples and is computationally negligible.
    
\item {\bf Code of ethics}
    \item[] Question: Does the research conducted in the paper conform, in every respect, with the NeurIPS Code of Ethics \url{https://neurips.cc/public/EthicsGuidelines}?
    \item[] Answer: \answerYes{} 
    \item[] Justification: The research uses only public benchmark datasets and conforms with the NeurIPS Code of Ethics.

\item {\bf Broader impacts}
    \item[] Question: Does the paper discuss both potential positive societal impacts and negative societal impacts of the work performed?
    \item[] Answer: \answerYes{} 
    \item[] Justification: Broader impacts are discussed in Section~\ref{sec:broader_impact}: improved auditability for regulated decisions, alongside risks of over-reliance on an LLM feature extractor and increased API cost.
    
\item {\bf Safeguards}
    \item[] Question: Does the paper describe safeguards that have been put in place for responsible release of data or models that have a high risk for misuse (e.g., pre-trained language models, image generators, or scraped datasets)?
    \item[] Answer: \answerNA{} 
    \item[] Justification: The paper releases no high-risk pretrained models or scraped datasets.

\item {\bf Licenses for existing assets}
    \item[] Question: Are the creators or original owners of assets (e.g., code, data, models), used in the paper, properly credited and are the license and terms of use explicitly mentioned and properly respected?
    \item[] Answer: \answerYes{} 
    \item[] Justification: All datasets and the LLM used are cited in Section~\ref{sec_results}; the datasets are standard, publicly available benchmarks.

\item {\bf New assets}
    \item[] Question: Are new assets introduced in the paper well documented and is the documentation provided alongside the assets?
    \item[] Answer: \answerNA{} 
    \item[] Justification: The paper does not release new datasets or models; evolved rubrics are reported in the appendix for illustration only.

\item {\bf Crowdsourcing and research with human subjects}
    \item[] Question: For crowdsourcing experiments and research with human subjects, does the paper include the full text of instructions given to participants and screenshots, if applicable, as well as details about compensation (if any)? 
    \item[] Answer: \answerNA{} 
    \item[] Justification: The paper does not involve crowdsourcing or research with human subjects.

\item {\bf Institutional review board (IRB) approvals or equivalent for research with human subjects}
    \item[] Question: Does the paper describe potential risks incurred by study participants, whether such risks were disclosed to the subjects, and whether Institutional Review Board (IRB) approvals (or an equivalent approval/review based on the requirements of your country or institution) were obtained?
    \item[] Answer: \answerNA{} 
    \item[] Justification: The paper does not involve research with human subjects.

\item {\bf Declaration of LLM usage}
    \item[] Question: Does the paper describe the usage of LLMs if it is an important, original, or non-standard component of the core methods in this research? Note that if the LLM is used only for writing, editing, or formatting purposes and does \emph{not} impact the core methodology, scientific rigor, or originality of the research, declaration is not required.
    \item[] Answer: \answerYes{} 
    \item[] Justification: LLMs (Claude Haiku 4.5) are a core component as the feature generator and evaluator, and are fully described in Section~\ref{sec_methods}.

\end{enumerate}

\end{document}